\documentclass[11pt]{article}

\usepackage[margin=1in]{geometry}
\usepackage{amsmath,amssymb,amsthm,mathtools,bm}
\usepackage{booktabs,array,tabularx}
\usepackage{graphicx}
\usepackage{algorithm}
\usepackage{algpseudocode}
\usepackage[dvipsnames]{xcolor}
\usepackage{enumitem}
\usepackage{caption}
\usepackage{pifont}
\usepackage{xurl}
\usepackage{microtype}
\usepackage{fancyhdr}
\usepackage{placeins}
\usepackage[colorlinks=true,linkcolor=blue!50!black,citecolor=blue!50!black,urlcolor=blue!50!black]{hyperref}

\setlist{itemsep=2pt,topsep=4pt}
\newtheorem{proposition}{Proposition}
\newtheorem{lemma}{Lemma}
\theoremstyle{remark}
\newtheorem{remark}{Remark}

\newcommand{\R}{\mathbb{R}}
\newcommand{\Ahat}{\hat{A}}
\newcommand{\Bhat}{\hat{B}}
\newcommand{\one}{\mathbf{1}}
\newcommand{\diag}{\operatorname{diag}}
\newcommand{\orth}{\operatorname{orth}}
\newcommand{\range}{\operatorname{range}}
\newcommand{\blkdiag}{\operatorname{blkdiag}}
\newcommand{\svd}{\operatorname{svd}}
\newcommand{\qr}{\operatorname{qr}}
\newcommand{\eigh}{\operatorname{eigh}}

\newcommand{\nB}{n_{B}}
\newcommand{\nw}{n_{w}}
\newcommand{\ness}{n_{\mathrm{ESS}}}
\newcommand{\kt}{k_{t}}
\newcommand{\Rose}{\textsc{Roseland}}
\newcommand{\iRose}{\textsc{iRoseland}}

\algrenewcommand\algorithmicrequire{\textbf{Input:}}
\algrenewcommand\algorithmicensure{\textbf{Output:}}

\fancypagestyle{coverpage}{
    \fancyhf{}

    \fancyfoot[L]{%
        \parbox[b]{0.85\textwidth}{%
            \footnotesize \textit{AI was used for copy editing and language polishing.}%
        }%
    }
    
    \fancyfoot[R]{\thepage}
}

\newcommand{\figfile}[2][width=\linewidth]{%
  \IfFileExists{#2}{\includegraphics[#1]{#2}}%
  {\fbox{\parbox{0.9\linewidth}{\centering\texttt{\detokenize{#2}} (not found)}}}}

\title{Incremental Roseland: Batch-wise Landmark Diffusion\\
via Kernelized Sequential Karhunen--Lo\`eve Update}
\author{Jeanie Schreiber}
\date{November 2026}

\begin{document}
\maketitle
\thispagestyle{coverpage}
\begin{abstract}
Landmark diffusion (\Rose{}) computes a diffusion-map-style spectral embedding
from the singular value decomposition of a degree-normalized data-to-landmark kernel matrix, at a cost far below that of full diffusion maps. It assumes,
however, that all data are available at once. We introduce
incremental \Rose{} (\iRose{}), which casts the landmark affinity as a
row-growing matrix  whose SVD can be updated with a kernelized Sequential
Karhunen--Lo\`eve (SKL) scheme. Degre correction is handled by a first order expansion about a reference mean kernel row and so that the landmark-side eigenbasis may be recovered from a single landmark-by-landmark eigendecomposition at each step. The presented method requires only a single pass over the data and runs in bounded-memory with state independent of the number of points processed. A forgetting factor lets the method track drifting distributions, and a sample-size-invariant embedding keeps outputs comparable across time. Experiments show agreement with \Rose{} to round-off on stationary data, and accuracy that exceeds equal-memory \Rose{} by an order of magnitude on long streams. On drifting data, we observe large gains from forgetting with a window
matched to the drift.
\end{abstract}

\tableofcontents
\section{Introduction}

Spectral embeddings such as diffusion maps~\cite{coifman2006} and Laplacian
eigenmaps~\cite{belkin2003} recover the geometry of a data manifold from the
leading eigenpairs of a graph operator built on the samples. However their high memory cost is an obstacle to large scale use, requiring $\mathcal{O}(n^2)$ memory and $\mathcal{O}(n^3)$ decomposition for a dense $n\times n$ kernel matrix. \Rose{} (RObust and Scalable Embedding via LANdmark Diffusion)~\cite{shen2022jmlr,shen2022ii}
reduces this cost by routing the diffusion through a small landmark set $Y=\{y_k\}_{k=1}^m$ with $m\ll n$. It
forms the thin $n\times m$ data-to-landmark cross kernel $W^{(r)}$, degree-normalizes it, and computes a singular
value decomposition (SVD). The resulting singular vectors converge to eigenfunctions of the
Laplace--Beltrami operator~\cite{shen2022jmlr,shen2022ii,singer2006,belkin2008} while remaining robust to noise. For $n$ data points in $\R^p$ and $m= n^\beta$ landmarks the cost is $\mathcal{O}(pn^{1+\beta}+n^{1+2\beta})$, comparable to Nystr\"om-type
methods~\cite{williams2001,fowlkes2004}.

Many scientific and sensor datasets are not available all at once. Long-term physiological recordings~\cite{goldberger2000}, bedside monitoring signals~\cite{johnson2016}, and wearable sensor streams~\cite{vijayan2021} arrive in batches, may be too large to store or re-read, and are most useful when the embedding is current after every batch~\cite{shen2022jmlr}. Refitting
\Rose{} on all data seen so far costs $O(nm^2)$ per batch and requires every kernel row to be
retained, so both latency and memory grow with the stream. What is wanted is an update whose
cost depends on the batch size and the landmark count, not on the amount of data already processed.

The Sequential Karhunen--Lo\`eve (SKL) algorithm of Levy and Lindenbaum~\cite{levy2000}, as extended
by Ross et al.~\cite{ross2008}, provides exactly this for principal component analysis by maintaining a truncated
SVD of a growing matrix and updating it from each new batch without revisiting old observations. The premise of this paper is that \Rose's central computational object, the SVD
of the degree-normalized affinity $\Ahat=D^{-1/2}W^{(r)}\in\R^{n\times m}$, has the same form as the matrix SKL
maintains. A new batch of data points appends rows to $\Ahat$ while the landmark dimension $m$ stays
fixed. Transposed, $\Ahat^\top$ grows by columns, which is precisely the SKL setting. The matrix being factored is a kernel (Gram-type) block rather than
raw data, so the scheme is a kernelized SKL update, and tracking its SVD amounts to tracking the
diffusion eigenbasis of a streaming manifold. We call the resulting method incremental \Rose{}
(\iRose).

The central difficulty in applying SKL to \Rose's degree normalized affinity matrix is that degree normalization
couples every row to all of the data. The degree of point $x_i$ is $d_i=w_i^\top c$, where $c=W^{(r)\top}\one\in\R^m$
is the landmark column sum. When a batch arrives, $c$ changes to $c+c_B$ and every stored degree
changes, so every old row of $\Ahat$ is rescaled by $\sqrt{d_i^{\mathrm{old}}/d_i^{\mathrm{new}}}\le 1$. In SKL, old observations never
change, so the standard update does not model this. A correct combination must therefore update the
old rows for the change in degree and apply any forgetting factor
consistently to the degrees as well as the spectrum. Any exact method must either retain per-point information growing with $n$, or re-read the data. \iRose{} instead replaces the per-point degree correction by a first-order expansion with $n$-independent state. We also fix a sample-size-invariant embedding scale and a gauge for the near-degenerate eigenspaces that arise on symmetric manifolds. Section~\ref{sec:combining} develops these requirements.

An exact version of \iRose{} maintains a direct correspondence with \Rose{} by rescaling and refactorizing every stored row. This requires the kernel row and left singular vector of every point (or alternatively, re-reading each batch), so its cumulative cost is no lower than that of \Rose{} on all of the data. The first-order correction used by the main version of \iRose{} instead expands the degrees about a reference mean kernel row and accumulates the statistics $T_0\in\R^{m\times m}$ and $T_1\in\R^{m\times m\times m}$, giving a true single-pass batching scheme whose state does not grow with the stream. The exact version serves as a correctness reference for small streams, and the first-order version targets long streams under a memory budget. We emphasize that SKL enters the first-order method only through the correspondence of Section~\ref{sec:combining}. Once all $m$ landmark directions are tracked, the SKL rotations collapse to a single eigendecomposition of an $m\times m$ matrix assembled from accumulated statistics, and no SKL update is performed explicitly.

Both versions maintain \Rose{} on data weighted
by age, such that a point that arrived $a$ batches ago has weight $f^{2a}$, where $f\in[0,1]$ is the forgetting factor. The choice of $f$ trades two
kinds of error. Sampling error falls like $1/\sqrt{\ness}$, where $\ness$ is the effective sample size of the geometric weights, and lag error arises from
describing the data as it was over the window rather than as it is now. It vanishes if the sampling distribution is fixed and grows with the window if the distribution drifts. We argue, and confirm empirically, that $f=1$ is the right choice for stationary sampling, and $f<1$ is best when the distribution drifts, with $f$ chosen so that the window spans a stretch of the stream over which the data are roughly unchanged (Section~\ref{sec:forgetting}).

With $f=1$ and full storage, exact \iRose{} and \Rose{} on the
same points are the same estimator, so \iRose{} offers no accuracy advantage on equal data. Its
advantages are in resources, \emph{i.e.} a bounded memory, single-pass operation without re-reading or storing
the stream, supplying current estimates after every batch at constant cost. With $f<1$ the method additionally adapts to a changing distribution, at the price of a permanent sampling-error
floor set by the effective sample size $\ness$. A chunked two-pass \Rose{} can also run in bounded memory, but it requires the data to be read twice and does not provide a running estimate. We also note that the landmark set is fixed in the present work, so drift in the landmark geometry is not tracked. Forgetting acts on the sample weights, and data that move away from the landmarks produce small, poorly conditioned degrees. Future work will address the case where landmarks are updated to more accurately reflect streaming data.

\paragraph{Contributions.}
We first establish the correspondence between \Rose{} and SKL
(Section~\ref{sec:combining}) and identify what must be added to SKL in order to
track a degree-normalized landmark affinity. To make embeddings from different times directly comparable we also describe a sample-size-invariant, gauge-fixed embedding. We then present the main \iRose{} algorithm based on a first-order degree expansion with periodic
re-referencing (Section~\ref{sec:algorithm}). For $f=1$, a heuristic argument gives a linearization error of $O(\log n/n)$. When measured, it stays within
$0.5$--$5\%$ of the $O(n^{-1/2})$ sampling error. The same construction
supports $f<1$ by weighting the accumulated statistics. Under drift its linearization error grows with the window length and the drift rate,
and is small at a well-chosen $f$ but not for windows that are too long. Finally, we provide numerical results on $\mathbb{S}^1$ and $\mathbb{S}^2$, comparing exact \iRose{} with \Rose{} on the same data, as well as an equal-memory scaling study where resources are constrained to the same $5.0$ MB budget. We also give results for data with drifting distribution to gauge the effects of the forgetting factor.



\section{Background}\label{sec:background}
Incremental spectral computations have been studied in several settings.
Brand~\cite{brand2006} and Baker et al.~\cite{baker2012} give low-rank updates
of the thin SVD, and incremental eigen-system updates have been developed for
spectral clustering of evolving data~\cite{ning2010,dhanjal2014}. Law and
Jain~\cite{law2006} extend manifold-learning embeddings to incremental data.
These approaches update a fixed matrix factorization or a graph, whereas the
degree normalization in \Rose{} rescales every previously processed row when a
batch arrives. Handling this coupling, and doing so under a bounded-memory
budget, is the main focus of the present work.

\subsection{Notation}

We follow the notation convention of \Rose{} (with the exception of Section~\ref{sec:skl}) and translate SKL into this
notation in Section~\ref{sec:combining}. The symbols below are used in the remainder of the paper.

\begin{center}
\small
\begin{tabular}{@{}ll@{}}
\toprule
Symbol & Meaning\\
\midrule
$x_i\in\R^p,\ i=1,\dots,n$ & data points seen so far\\
$y_k,\ k=1,\dots,m$ & fixed landmarks, $m\ll n$\\
$K_\epsilon(x,y)=e^{-\|x-y\|^2/\epsilon}$ & Gaussian kernel with bandwidth $\epsilon$\\
$w_i\in\R^m,\ (w_i)_k=K_\epsilon(x_i,y_k)$ & kernel row of $x_i$\\
$W\in\R^{n\times m}$ & landmark-set affinity with rows $w_i^\top$ (denoted $W^{(r)}$ in~\cite{shen2022jmlr})\\
$W_B\in\R^{\nB\times m},\ \nB$ & kernel rows of an incoming batch; batch size\\
$c=W^\top\one\in\R^m$ & column sum of $W$; $c_B=W_B^\top\one$ for a batch\\
$d_i=w_i^\top c,\ D=\diag(d)$ & \Rose{} degrees\\
$\Ahat=D^{-1/2}W=U\Lambda V^\top$ & normalized affinity and its SVD; $U\in\R^{n\times k}$, $V\in\R^{m\times k}$\\
$\Lambda=\diag(\sigma_1,\dots,\sigma_k)$ & singular values, $\sigma_1\ge\sigma_2\ge\cdots$\\
$\Bhat=\diag(W_Bc)^{-1/2}W_B$ & degree-normalized batch rows\\
$q,\ t$ & embedding dimension; diffusion time\\
$f\in[0,1]$ & forgetting factor\\
$\omega_i,\ \Omega=\diag(\omega)$ & sample weights, $\omega_i=f^{2a_i}$ with $a_i$ the number of updates since $x_i$ arrived\\
$\nw=\sum_i\omega_i$ & weighted sample count (effective window); $\nw=n$ when $f=1$\\
$\ness=(\sum_i\omega_i)^2/\sum_i\omega_i^2$ & effective sample size of the weights; $\ness=n$ when $f=1$\\
$M_{a:b}$ & columns $a$ through $b$ of a matrix $M$ (1-based)\\
$\mathcal{G}$ & orthogonal gauge (Procrustes) matrix, Section~\ref{subsec:embedding}\\
$\Delta,\ \tau$ & drift per batch; batch index in the drift study\\
$\qr(M)$ & economy QR factorization\\
\bottomrule
\end{tabular}
\end{center}

\vspace{2mm}

\noindent Two identities are used repeatedly. The right Gram matrix of $\Ahat$ is
\begin{equation}\label{eq:gram}
G:=\Ahat^\top\Ahat=\sum_{i=1}^n\frac{w_iw_i^\top}{w_i^\top c}\in\R^{m\times m},
\end{equation}
so that $V$ and $\Lambda^2$ are its eigenvectors and eigenvalues. Because $\Ahat=U\Lambda V^\top$, the left singular vector
of any point with kernel row $w$ extends out of sample through
\begin{equation}\label{eq:oos}
u(x)^\top=\frac{w^\top}{\sqrt{w^\top c}}\,V\Lambda^{-1}.
\end{equation}

\subsection{Roseland}\label{sec:roseland}

\Rose{} (RObust and Scalable Embedding via LANdmark Diffusion)~\cite{shen2022jmlr} builds a diffusion-map-like
embedding without forming an $n\times n$ affinity. Instead of connecting data points directly, it connects
each data point to a small landmark set $Y$ and lets diffusion pass through the landmarks. The
induced $n\times n$ affinity is $WW^\top$, and $D^{-1}WW^\top$ is a row-stochastic transition matrix on the data,
which is never formed explicitly.

The algorithm proceeds as follows. First, compute the data-to-landmark cross kernel matrix $W\in\R^{n\times m}$ with entries $W_{ik}=K(x_i,y_k)$. Next,
form the column sum $c=W^\top1$ and the degrees $d=Wc$, so that
$D_{ii}=e_i^\top WW^\top1$. Then compute the thin SVD of the
degree-normalized cross kernel, $D^{-1/2}W=U\Lambda V^\top$. The left singular
vectors are rescaled as $\tilde U=D^{-1/2}U$. Discard the first column,
which is the trivial constant mode, and keep columns $2,\dots,q+1$, with $\Lambda_{2:q+1}=\diag(\sigma_2,\dots,\sigma_{q+1})$. Finally, each point
is embedded as
$\Phi_t(x_i)=e_i^\top\tilde U_{2:q+1}\Lambda_{2:q+1}^{2t}$, where $t$ is the
diffusion time.

As $n\to\infty$ the singular vectors converge to eigenfunctions of the Laplace--Beltrami operator on
the underlying manifold~\cite{shen2022jmlr,shen2022ii}. With $m= n^\beta$ landmarks the cost is $\mathcal{O}(pn^{1+\beta}+n^{1+2\beta})$, substantially
below the cost of full diffusion maps, and the choice $\beta<1/2$ is recommended.

\begin{algorithm}[h!]
\caption{\Rose}\label{alg:roseland}
\begin{algorithmic}[1]
\Require data $X=\{x_i\}_{i=1}^n$; landmarks $Y=\{y_k\}_{k=1}^m$; bandwidth $\epsilon$; embedding dimension $q$; diffusion time $t$
\Ensure embedding $\Phi\in\R^{n\times q}$
\State $W_{ik}\gets K_\epsilon(x_i,y_k)$ 
\State $c\gets W^\top\one$;\quad $d\gets Wc$ 
\State $[U,\Lambda,V]\gets\svd\bigl(\diag(d)^{-1/2}W\bigr)$ 
\State $\bar U\gets\diag(d)^{-1/2}U$ 
\State $\Phi\gets\bar U_{2:q+1}\,\Lambda_{2:q+1}^{2t}$ 
\end{algorithmic}
\end{algorithm}

The following lemma shows that age-weighted \Rose{} is well-posed. This is the form that the forgetting factor produces. It holds for any positive diagonal weighting $\Omega$, and \Rose{} is the case $\Omega=I$.

\begin{lemma}\label{lem:trivial}
Let $\Omega$ be any positive diagonal weight matrix and let $W$ be entrywise positive, as for the Gaussian kernel. Define the weighted
column sum $c=W^\top\Omega\one$, the degrees $D=\diag(Wc)$, and $\Ahat=\Omega^{1/2}D^{-1/2}W$. Then $\sigma_1(\Ahat)=1$, this singular value is simple, and
the corresponding right singular vector is $v_1=c/\|c\|$ (unique up to sign).
\end{lemma}
\begin{proof}
Let $X=\Omega^{1/2}D^{1/2}$. Then $X^{-1}(\Ahat\Ahat^\top)X=D^{-1}WW^\top\Omega$. This matrix is entrywise nonnegative
with row sums $D^{-1}W(W^\top\Omega\one)=D^{-1}Wc=\one$, so its spectral radius is 1. This is the standard argument that the normalized affinity of a nonnegative
weighted graph has leading singular value one~\cite{chung1997}. It is similar to the
positive semidefinite matrix $\Ahat\Ahat^\top$, whose eigenvalues are the $\sigma_j^2$, hence $\sigma_1^2=1$. The left singular
vector is $u_1\propto X\one=\Omega^{1/2}D^{1/2}\one$, and $\Ahat^\top u_1\propto W^\top\Omega\one=c$. Since $W$ is entrywise positive, $D^{-1}WW^\top\Omega$ is a positive matrix, so by Perron--Frobenius~\cite{pillai2005perron} the eigenvalue $1$ is simple, which gives simplicity of $\sigma_1$ and uniqueness of $v_1$ up to sign.
\end{proof}

\subsection{Sequential Karhunen--Lo\`eve}\label{sec:skl}

SKL maintains a truncated SVD of a matrix that grows by columns, without retaining or revisiting
the earlier columns. In this subsection only, we follow the presentation of Ross et al.~\cite{ross2008}: observations
are stored as columns, $A\in\R^{p\times N}$ has the rank-$k$ truncated SVD $A\approx U_A\Sigma V_A^\top$, and $B\in\R^{p\times\ell}$ is a
batch of $\ell$ new columns. Then
\begin{equation}\label{eq:skl}
[A\ \ B]=\bigl[U_A\ \ \tilde B\bigr]\,R\begin{bmatrix}V_A^\top&0\\0&I\end{bmatrix},\qquad
R=\begin{bmatrix}\Sigma&U_A^\top B\\0&\tilde B^\top\bigl(B-U_AU_A^\top B\bigr)\end{bmatrix},
\end{equation}
where $\tilde B=\orth(B-U_AU_A^\top B)$ is an orthonormal basis for the component of $B$ orthogonal to $U_A$.

The update proceeds in three steps. First, compute the orthonormal basis
$\tilde B$ of the residual $B-U_A U_A^\top B$ and form the small core matrix $R$
of~\eqref{eq:skl}. Second, compute the SVD of the
small core, $R=\tilde U\tilde\Sigma\tilde V^\top$. Third, rotate the basis, $U_A\leftarrow[\,U_A\;\tilde B\,]\tilde U$, set
$\Sigma\leftarrow\tilde\Sigma$, and discard the components beyond rank $k$. The matrix $R$ has size $(k+r)\times(k+\ell)$, where $r\le\ell$ is the rank of the residual, so the
cost per update depends on $p$, $k$ and $\ell$ but not on the number $N$ of columns already processed.

\begin{algorithm}[h!]
\caption{SKL update with forgetting (notation of~\cite{ross2008})}\label{alg:skl}
\begin{algorithmic}[1]
\Require current factors $U_A\in\R^{p\times k}$, $\Sigma\in\R^{k\times k}$; new columns $B\in\R^{p\times\ell}$; forgetting factor $f$; rank $k$
\Ensure updated $U_A'$, $\Sigma'$
\State $\tilde B\gets\orth\bigl(B-U_AU_A^\top B\bigr)$ 
\State $R\gets\begin{bmatrix}f\Sigma&U_A^\top B\\0&\tilde B^\top(B-U_AU_A^\top B)\end{bmatrix}$ \Comment{ forgetting enters here}
\State $[\tilde U,\tilde\Sigma,\sim]\gets\svd(R)$
\State $U_A'\gets[U_A\ \ \tilde B]\,\tilde U_{:,1:k}$;\quad $\Sigma'\gets\tilde\Sigma_{1:k}$
\end{algorithmic}
\end{algorithm}

Ross et al.\ add a forgetting factor $f\in[0,1]$ by replacing $\Sigma$ with $f\Sigma$ in $R$,
which is equivalent to factorizing $[fU_A\Sigma\ \ B]$. Each update
multiplies the amplitude of all earlier columns by $f$, so a block added $j$ updates ago contributes
to the modelled covariance with weight $f^{2j}$. With $f=1$ all blocks count equally, which is the stationary setting. With $f<1$ the decomposition follows recent data and gradually discards old data, which is useful when the sampling distribution changes. Their running count of the sample
history, used in the mean update, evolves as $n\gets fn+\ell$ and equilibrates at $\ell/(1-f)$.

\begin{remark}\label{rem:window}
Three points to note for \Rose:
\begin{enumerate}
\item The covariance (spectral) weight of old data is $f^{2j}$, whereas the mean-history count uses $f$. The
spectral weight mass, or window, is therefore $\nw=\ell/(1-f^2)$, not $\ell/(1-f)$.
\item Because the weights are unequal, the number of equally weighted
points with the same variance is the Kish effective sample size~\cite{kish1965}
\[
\ness=\frac{(\sum_i\omega_i)^2}{\sum_i\omega_i^2}=\ell\,\frac{1+f^2}{1-f^2}=(1+f^2)\,\nw,\approx2n_w
\]

and estimates have sampling error of order $\ness^{-1/2}$, so a shorter window is noisier. Its benefit is a smaller lag
behind a changing distribution. The window $\nw$ sets the time scale of the lag, while $\ness$ sets the variance. The trade-off between sampling error and lag error is the familiar one in
adaptive windowing for non-stationary streams~\cite{gama2014}. Section~\ref{sec:forgetting} makes this trade-off explicit and gives guidance for choosing $f$.
\item \Rose{} does not center the data, so the mean-correction variant of Ross et al.\ does not apply.
Only plain SKL and the forgetting factor are used.
\end{enumerate}
\end{remark}

\vspace{2mm}

\noindent Both algorithms revolve around a thin SVD, and \Rose's SVD of $D^{-1/2}W$ is the quantity
from which the embedding is read off. In SKL, appending columns to $A$ leaves the existing columns unchanged. In \Rose, however, appending rows to $W$ changes $c$ and therefore every degree $d_i=w_i^\top c$,
so the existing rows of $\Ahat$ are rescaled. If old
data are down-weighted using the forgetting factor, they must be down-weighted in the degrees as well as in the spectrum, so that the result is \Rose{} on age-weighted data (Lemma~\ref{lem:trivial}) rather than an inconsistent mixture. This degree coupling, together with its consequences for forgetting, is the subject of Section~\ref{sec:combining}.

\section{Combining Roseland and SKL}\label{sec:combining}

This section explains how \Rose{} is cast as a row-growing SKL problem and explains where the two algorithms
conflict, as well as which modifications are needed for the combination to track the \Rose{}
SVD exactly. We then discuss the forgetting factor
and when it should be used. Section~\ref{sec:algorithm} assembles these pieces into the full algorithms.

Throughout, $\kt$ denotes the number of singular triplets \emph{tracked} by the state, and $k\le\kt$ the
number reported. The state after $n$ points is $(U,\Lambda,V)$ with $U\in\R^{n\times\kt}$, $\Lambda\in\R^{\kt\times\kt}$ and $V\in\R^{m\times\kt}$,
such that $U\Lambda V^\top$ approximates $\Ahat=D^{-1/2}W$. We also write $Q_B=\Bhat^\top\in\R^{m\times\nB}$ for the
degree-normalized batch stored as columns.

\subsection{The correspondence}

A new batch appends $\nB$ rows to $\Ahat\in\R^{n\times m}$ while the landmark dimension $m$ stays fixed. Transposing,
$\Ahat^\top\in\R^{m\times n}$ gains columns, which is the setting of Section~\ref{sec:skl}. The tracked basis is therefore
the landmark-side factor $V$, and SKL's never-formed right factor is \Rose's left factor $U$.

\begin{center}
\small
\begin{tabular}{@{}p{0.42\linewidth}p{0.5\linewidth}@{}}
\toprule
SKL (Ross et al.) & Row-growing \Rose{} (this paper)\\
\midrule
Data matrix $A$, observations as columns & $\Ahat^\top\in\R^{m\times n}$\\
Tracked basis $U_A$ and values $\Sigma$ & Landmark-side $V$ and $\Lambda$\\
New columns $B$ & $Q_B=\Bhat^\top\in\R^{m\times\nB}$\\
Ambient dimension $p$ (fixed) & Number of landmarks $m$ (fixed)\\
Number of observations (growing) & Number of data points $n$ (growing)\\
Batch size $\ell$ & $\nB$\\
Retained rank $k$ & Tracked rank $\kt$\\
Residual basis $\tilde B$ & $V_p\in\R^{m\times r}$\\
$[\tilde U,\tilde\Sigma,\tilde V]$ from the SVD of $R$ & $[U_R,\Sigma_R,Z_R]$\\
\bottomrule
\end{tabular}
\end{center}

The matrix being factored is a kernel block rather than raw data, so the scheme is a kernelized
SKL, \emph{i.e.} each row of $\Ahat$ is the degree-normalized vector of kernel evaluations $K_\epsilon(x_i,\cdot)$ against the
landmarks. In this notation, identity~\eqref{eq:skl} reads
\begin{equation}\label{eq:skl-rose}
\begin{bmatrix}\Ahat\\\Bhat\end{bmatrix}^\top=\bigl[V\Lambda U^\top\ \ Q_B\bigr]=\bigl[V\ \ V_p\bigr]\,R\begin{bmatrix}U^\top&0\\0&I\end{bmatrix},\qquad
R=\begin{bmatrix}\Lambda&V^\top Q_B\\0&V_p^\top\bigl(Q_B-VV^\top Q_B\bigr)\end{bmatrix},
\end{equation}
where $V_p$ is an orthonormal basis of the residual $Q_B-VV^\top Q_B$. Writing the SVD $R=U_R\Sigma_R Z_R^\top$,
the stacked matrix has
\begin{equation}\label{eq:skl-update}
V'=[V\ \ V_p]\,U_R,\qquad\Lambda'=\Sigma_R,\qquad U'=\blkdiag(U,I_{\nB})\,Z_R.
\end{equation}
Because $V$ has $m$ rows, the SVD of $R$ has size $(\kt+r)\times(\kt+\nB)$ with $r\le m-\kt$, which does
not depend on $n$. Since $\Ahat^\top\Ahat=G$ in~\eqref{eq:gram}, tracking $(V,\Lambda)$ determines the eigenpairs of
the transition operator $D^{-1}WW^\top$ on the data, that is, the diffusion eigenbasis of the stream.

\subsection{The degree coupling}\label{sec:coupling}

The correspondence above would be complete if the old rows of $\Ahat$ were unchanged by new data. In \Rose{} however, the rows of $\hat A$ are normalized by a degree that depends on all of the data. A batch updates the column sum to $c\gets c+c_B$ with $c_B=W_B^\top\one$, and since $d_i=w_i^\top c$,
the degree of every stored point changes:
\begin{equation}\label{eq:deg-update}
d_i^{\mathrm{new}}=d_i^{\mathrm{old}}+w_i^\top c_B,\qquad i\le n.
\end{equation}
The old rows of $\Ahat$ are therefore rescaled by
\begin{equation}\label{eq:scale}
s_i=\sqrt{d_i^{\mathrm{old}}/d_i^{\mathrm{new}}},
\end{equation}
so the old block of the updated matrix is $\diag(s)\,U\Lambda V^\top$, not $U\Lambda V^\top$. In SKL, old observations
never change, and the update~\eqref{eq:skl-update} does not model this rescaling.

Handling the old-row rescaling needs per-row information. This leaves a choice of how to carry out the correction. To set up a direct correspondence with \Rose, we can carry out the correction exactly at the cost of storing every kernel row and left singular vector. Doing so requires three ingredients beyond the plain SKL update, which we summarize here and discuss in more detail in Appendix~\ref{sec:appendix}.

To carry out the exact degree correction, old rows are rescaled by $s_i=\sqrt{d_i^{\mathrm{old}}/d_i^{\mathrm{new}}}$, the rescaled left factor $\mathrm{diag}(s)U$ is re-orthogonalized by a thin QR
factorization, and the resulting small $\kt\times\kt$ core is refactorized by an
SVD. After this step
$U\Lambda V^\top=\mathrm{diag}(s)\,U_{\mathrm{old}}\Lambda_{\mathrm{old}}V_{\mathrm{old}}^\top$
holds exactly with orthonormal $U$ and $V$, however two further considerations must
accompany this step if exactness is to be preserved. First, because the degree correction revisits the stored left factor at every update, $U$ must be carried along and rotated by the SKL update. Without this bookkeeping orthonormality is not maintained, and the next QR counts the previous batch's energy a second
time. Second, exactness requires tracking all $\kt=m$ triplets, so that no singular
value is truncated before the correction acts.  The SKL core matrix requires the residual basis $V_p$ to be orthonormal and orthogonal to $V$ as in~\ref{eq:skl-rose}. A plain economy QR of the incoming batch overlaps $\range(V)$ and double-counts the in-subspace part of every batch, so $V_p$ is instead built from a rank-revealing SVD of the projected residual (Appendix~\ref{sec:appendix}). With these in place, the update maintains the SVD of the (age-weighted) \Rose{} matrix after every batch
(Proposition~\ref{prop:exact}).

The exact procedure allows us to carry out a batched version of \Rose{} directly, but requires storing $W$ and $U$ for every point. Its memory is $O(n\cdot 2m)$
and each update costs $O(nm^2+\nB m^2)$ with $\kt=m$, dominated by the QR.
Cumulatively this is no cheaper than refitting plain \Rose. Its value lies in its exactness, serving as the reference against which approximations are checked. (The comparison that tests the exact scheme against plain \Rose{} on the same weighted data is made in Section~\ref{sec:correctness}.) It is practical only for small streams, while longer streams require removing the per-row dependence. This can be done to first order by accumulating aggregate statistics of the data about a reference mean kernel row, with memory independent of $n$. This procedure and its algorithm are developed in Section~\ref{sec:algorithm}.

\subsection{The forgetting factor: stationary versus drifting data}\label{sec:forgetting}

Ross et al.\ apply the forgetting factor to
the spectrum, $\Lambda\gets f\Lambda$. In \Rose, what this represents is a weighting of the data by age,
and the degrees and column sum must be weighted the same way. Suppose point $x_i$ carries weight
$\omega_i=f^{2a_i}$, where $a_i$ counts the updates since it arrived, and define the weighted \Rose{} matrix
\begin{equation}\label{eq:weighted}
\Ahat=\Omega^{1/2}D^{-1/2}W,\qquad c=W^\top\Omega\one,\qquad D=\diag(Wc).
\end{equation}
At a batch update, old weights become $\omega_i\gets f^2\omega_i$ and new points get weight one, so
\begin{equation}\label{eq:weighted-update}
c\gets f^2c+c_B,\qquad d_i\gets f^2d_i^{\mathrm{old}}+w_i^\top c_B,
\end{equation}
and each old row $\sqrt{\omega_i}\,w_i^\top/\sqrt{d_i}$ of $\Ahat$ is multiplied by $f s_i$, with $s_i$ as in~\eqref{eq:scale} computed from the new degrees. The factor $f$ is Ross's forgetting applied to $\Lambda$, and the factor $s_i$ is the degree correction. Since
$d_i^{\mathrm{new}}\ge f^2d_i^{\mathrm{old}}$, the product satisfies $fs_i\le1$ for all $f$. The two factors must use the same weights, so forgetting has to enter the degrees through~\eqref{eq:weighted-update}.

The factor $f$ sets the effective window
\begin{equation}\label{eq:window}
\nw=\sum_i\omega_i\;\longrightarrow\;\frac{\nB}{1-f^2}
\end{equation}
of points that really count. For example, $f=0.95$ with batches of 5{,}000 points gives a window of
about $5.1\times10^4$ points. The choice of $f$ is a bias--variance trade-off between two kinds of error.
\begin{itemize}
\item \emph{Sampling error} falls like $1/\sqrt{\ness}$, so a smaller window is noisier. Because the geometric
weights are unequal, it is the error of a \Rose{} fit to $\ness=(1+f^2)\nw$ (about $2\nw$ for $f$ near one) equally weighted points
(Remark~\ref{rem:window}), not to $\nw$ points.
\item \emph{Lag error} comes from describing the data as it was over the window rather than as it is now. It
is zero if the sampling distribution never changes and grows with the window if it drifts. As a
heuristic, the age of the weighted data is geometrically distributed with ratio $f^2$, so its mean is
$f^2/(1-f^2)$ batches and its standard deviation is $f/(1-f^2)$ batches. A density moving by $\Delta$
per batch is therefore described by a mixture that is both \emph{shifted}, by about $\Delta f^2/(1-f^2)$, and
\emph{broadened}, by about $\Delta f/(1-f^2)$.
\end{itemize}

When the sampling distribution is stationary, one should use $f=1$ so that every point counts equally. In that case there is no lag error and the sampling error decreases as $n^{-1/2}$ without a floor. Using $f<1$ on stationary data only discards information, so the error against \Rose{} on all data grows as $f$
decreases. When the sampling distribution drifts, one should instead use
$f<1$, chosen so that the window $\nw\approx\nB/(1-f^2)$ spans the stretch of
the stream over which the data are roughly unchanged. One approach is to tune $f$ on held-out recent batches. Section~\ref{sec:experiments} quantifies the trade-off on stationary and drifting streams.

\subsection{Embedding-level considerations}
\label{subsec:embedding}

Because $U$ has orthonormal columns over $n$ points, its
entries are $O(n^{-1/2})$, and an embedding read from $U\Lambda^{2t}$ shrinks as data accumulate. We therefore embed with
\begin{equation}\label{eq:embed}
\Phi(x)=\sqrt{\nw}\,\frac{w(x)^\top}{\sqrt{w(x)^\top c}}\,V_{2:q+1}\Lambda_{2:q+1}^{2t-1}=\sqrt{\nw}\,u(x)^\top_{2:q+1}\Lambda^{2t}_{2:q+1},
\end{equation}
with $\nw=\sum_i\omega_i$ ($\nw=n$ for plain \Rose{} with $f=1$). Formula~\eqref{eq:oos} gives the left vector of a point of weight one. In the weighted case the in-sample row of an old point is $\sqrt{\omega_i}\,u(x_i)^\top$, so $\sum_iu_{ij}^2=1$ reads $\sum_i\omega_i\,u_j(x_i)^2=1$, and the factor
$\sqrt{\nw}$ rescales to eigenfunctions of unit norm in the $\omega$-weighted empirical measure (the empirical $L^2(p)$ norm when $f=1$), so the scale is $O(1)$ for every $n$. Equation~\eqref{eq:embed} is the symmetric form based
on $U$, while \Rose's published embedding uses $\bar U=D^{-1/2}U$, and a scale-invariant version of that
form is $\Phi(x)=\bigl(w^\top/(w^\top\mu)\bigr)V\Lambda^{2t-1}$ with $\mu=c/\nw$. The two forms are related by the point-dependent factor $(w^\top\mu)^{-1/2}$, the inverse square root of the normalized degree, so they agree up to scale only when degrees are nearly constant. Therefore the two differ only by a constant factor for uniform sampling (as with equispaced landmarks on $\mathbb{S}^1$), but for nonuniform densities they differ.

On symmetric manifolds such as $\mathbb{S}^1$ the population eigenvalues come in
exactly degenerate groups, and with finitely many samples the groups are split only by $O(n^{-1/2})$.
Each SVD may therefore return a degenerate pair rotated by an arbitrary angle, so that embeddings
from different updates are mutually rotated and no single global Procrustes alignment applied after the fact to the concatenated embedding can undo
this, because the rotation differs from batch to batch. In order to compare embeddings across updates, we compute the orthogonal map
\begin{equation}\label{eq:gauge}
\mathcal{G}=\operatorname*{arg\,min}_{\mathcal{G}^\top \mathcal{G}=I}\bigl\|V_{2:q+1}\mathcal{G}-V_{\mathrm{frame}}\bigr\|_F,
\end{equation}
which is orthogonal Procrustes~\cite{schonemann1966} with solution $\mathcal{G}=U_\mathcal{G}V_\mathcal{G}^\top$ from the SVD $V_{2:q+1}^\top V_{\mathrm{frame}}=U_\mathcal{G}\Sigma_\mathcal{G}V_\mathcal{G}^\top$~\cite{golub2013}.
The frame is initialized as $V_{\mathrm{frame}}\gets V_{2:q+1}$ after the first batch. We then rotate the embedding coordinates by $\mathcal{G}$, and set $V_{\mathrm{frame}}\gets V_{2:q+1}\mathcal{G}$. Since $\mathcal{G}$ is applied after the
eigenvalue weighting, it is an isometry of the embedding. Note that $\mathcal{G}$ is a full $q\times q$ orthogonal matrix, not a block-diagonal one per degenerate group. When the eigenvalues within a group are split, as in the drift study, $\mathcal{G}$ may mix modes with different eigenvalues, so that $V_{2:q+1}\mathcal{G}$ is no longer an eigenbasis. The rotated embedding nonetheless remains an isometric copy of the unrotated one. A block-diagonal variant restricted to near-degenerate groups would avoid this mixing. Related to this, the number of reported
modes $k$ should not split a degenerate group.

\section{The iRoseland Algorithm}\label{sec:algorithm}

The exact degree correction of Section~\ref{sec:coupling} makes the tracked
decomposition identical to \Rose{} on (age-weighted) data, and we give it in
Appendix~\ref{sec:appendix} as a correctness reference. It is not practical for
large datasets or long streams since it must retain the kernel row and left singular vector of every point seen so far. The algorithm we propose, which we simply call iRoseland, removes this per-row dependence by correcting the degrees to first order in aggregate statistics. Its state is independent of the number of points processed, and it requires a single pass over the data.

\subsection{First-order degree correction with bounded memory}\label{sec:firstorder}

With $\kt=m$, the tracked state $(V,\Lambda)$ is the eigendecomposition of the $m\times m$ matrix
$G=\Ahat^\top\Ahat$ of~\eqref{eq:gram}. For the age-weighted target~\eqref{eq:weighted}, write $\nw=\sum_i\omega_i$ and $\mu=c/\nw$ for the
weighted mean kernel row. Then
\begin{equation}\label{eq:gmu}
G(\mu)=\sum_i\frac{\omega_iw_iw_i^\top}{w_i^\top c}=\frac1{\nw}\sum_i\frac{\omega_iw_iw_i^\top}{w_i^\top\mu},
\end{equation}
which for $f=1$ is $\omega_i=1$ and $\nw=n$. The only obstacle to a single pass is the degree $w_i^\top\mu$, which
depends on all the data and so changes for every old point as $\mu$ evolves. Instead of correcting stored
rows, we expand about a fixed reference $\mu_r$. Define $\delta = \mu-\mu_r$, with $g_i=w_i^\top\mu_r>0$ the fixed reference degree. Then
\begin{equation}\label{eq:eps}
w_i^\top\mu=g_i(1+\varepsilon_i),\qquad
\varepsilon_i=\frac{w_i^\top\delta}{w_i^\top\mu_r},
\end{equation}
so $\varepsilon_i$ is the relative change in the degree of $x_i$ when the mean kernel row moves from $\mu_r$ to $\mu$. Since all kernel entries are positive, we have $1+\varepsilon_i=w_i^\top\mu/w_i^\top\mu_r>0$ and the identity $1/(1+\varepsilon_i)=1-\varepsilon_i+\varepsilon_i^2/(1+\varepsilon_i)$ gives
\begin{equation}\label{eq:expand}
\frac1{w_i^\top\mu}=\frac{1}{g_i}\cdot\frac{1}{1+\varepsilon_i}=\frac1{g_i}-\frac{w_i^\top\delta}{g_i^2}+\frac1{g_i}\,\frac{\varepsilon_i^2}{1+\varepsilon_i}.
\end{equation}

\noindent Multiplying~\eqref{eq:expand} by $\omega_iw_iw_i^\top/\nw$ and summing,
\begin{equation}\label{eq:gsplit}
G(\mu)=\frac1{\nw}\Bigl[\sum_i\frac{\omega_iw_iw_i^\top}{g_i}
-\sum_i\frac{\omega_i(w_i^\top\delta)\,w_iw_i^\top}{g_i^2}
+\sum_i\frac{\omega_iw_iw_i^\top}{g_i}\frac{\varepsilon_i^2}{1+\varepsilon_i}\Bigr].
\end{equation}
The first sum is independent of $\delta$ and defines $T_0$. In the second, $w_i^\top\delta=\sum_cw_{ic}\delta_c$
is linear in $\delta$, so exchanging the sums over $i$ and $c$ gives, entrywise,
\[
\sum_i\frac{\omega_iw_{ia}w_{ib}(w_i^\top\delta)}{g_i^2}
=\sum_c\Bigl(\sum_i\frac{\omega_iw_{ia}w_{ib}w_{ic}}{g_i^2}\Bigr)\delta_c=(T_1\cdot\delta)_{ab},
\]
with
\begin{equation}\label{eq:T}
T_0=\sum_i\frac{\omega_iw_iw_i^\top}{w_i^\top\mu_r}\in\R^{m\times m},\qquad
(T_1)_{abc}=\sum_i\frac{\omega_iw_{ia}w_{ib}w_{ic}}{(w_i^\top\mu_r)^2}\in\R^{m\times m\times m},
\end{equation}
and $(T_1\cdot\delta)_{ab}=\sum_c(T_1)_{abc}\delta_c$. Dropping the last $\mathcal{O}(\varepsilon_i^2)$ sum in~\eqref{eq:gsplit},
\begin{equation}\label{eq:G1}
G(\mu)\approx\frac1{\nw}\bigl(T_0-T_1\cdot\delta\bigr).
\end{equation}
Let $\varepsilon=\max_i|\varepsilon_i|$ and assume $\varepsilon<1$. Each term of the dropped sum is positive semidefinite with coefficient in $[0,\varepsilon^2/(1-\varepsilon)]$, so
\[
0\preceq G(\mu)-\tfrac1{\nw}\bigl(T_0-T_1\cdot\delta\bigr)\preceq\frac{\varepsilon^2}{1-\varepsilon}\,\frac{T_0}{\nw}.
\]
Hence the first-order Gram matrix never overestimates $G$ and has relative error at most $\varepsilon^2/(1-\varepsilon)$ with respect to $T_0/\nw$.

The sums $T_0$ and $T_1$ depend on the data only through the fixed $\mu_r$, so each batch can be added to them
and then discarded. Since every old weight is multiplied
by $f^2$ at each update, $T_0$ and $T_1$ are multiplied by $f^2$ before the new batch is added, and likewise
$c\gets f^2c+c_B$ and $\nw\gets f^2\nw+\nB$. Every point, old or new, then receives its current degree to first
order, so there is no systematic asymmetry between old and new points in the degree correction.

The expansion degrades as $\mu$ moves away from $\mu_r$. Re-referencing moves the
reference to the current mean. Observe that $\nabla_\nu(w^\top\nu)^{-1} = -w/(w^\top\nu)^2$, so
\[DT_0(\mu_r)[\delta] = -\sum_i\frac{\omega_i(w_i^\top\delta)w_iw_i^\top}{(w_i^\top\mu_r)^2}=-T_1(\mu_r)\cdot\delta.\] Thus a first order Taylor expansion of $T_0(\cdot)$ from $\mu_r$ to $\mu$ is
\[T_0(\mu)\approx T_0(\mu_r)-T_1(\mu_r)\cdot(\mu-\mu_r),\] and we re-reference as
\begin{equation}\label{eq:rebase}
T_0\gets T_0-T_1\cdot(\mu-\mu_r),\qquad\mu_r\gets\mu,
\end{equation}
at a cost of $O(m^3)$. Two approximations are made at each re-reference: the second-order remainder of
the expansion for all accumulated points is dropped permanently, and $T_1$ is not moved to the new
reference (doing so would need the next-order tensor). This gives a relative error of $O(\varepsilon)$ in $T_1$, which later multiplies the next $\delta$, so its effect on $G$ is again second order.

The re-referencing schedule depends on the forgetting factor, because $f$
determines whether the weighted count $\nw$ keeps growing. For stationary data
with $f=1$, we re-reference whenever $n$ doubles. Heuristically, each reference
epoch $e$ bakes in an $O(\varepsilon_e^2)$ error for its share $n_e/n$ of the
points, and the stale $T_1$ adds a term of the same order. Because
$\varepsilon_e^2\sim 1/n_e$, each epoch contributes $O(1/n)$, and the total
linearization error over the $O(\log n)$ epochs is $O(\log n/n)$, which is
asymptotically negligible against the $O(n^{-1/2})$ sampling error. This is an
argument and not a proof. It predicts a local log--log slope near $-1$ (about $-0.9$ over $10^4\le n\le10^6$ once the logarithm is included), whereas the measured slopes in Section~\ref{sec:experiments} are $-0.77$ to $-0.84$. We attribute the gap to the stale $T_1$ term, but we have not verified this separately.

For $f<1$ the
weighted count stops growing, so a doubling rule would never trigger again, and
we instead re-reference at every batch, which costs only an $O(m^3)$ operation. On stationary data the step between consecutive references is $\delta\approx(\nB/\nw)(\mu_B-\mu_r)$, where $\mu_B$ is the batch mean kernel row. Since $\mu_B-\mu_r=O(\nB^{-1/2})$, this gives $\delta=O(\sqrt{\nB}/\nw)$ and $\varepsilon^2=O(\nB/\nw^2)$ per re-reference. A point remains in the window for about $\nw/\nB$ re-references, so the accumulated relative error in $G$ is of order
\[
\frac{\nw}{\nB}\cdot\frac{\nB}{\nw^2}=\frac1{\nw}=\frac{1-f^2}{\nB},
\]
which is $2\times10^{-5}$ at $f=0.95$ and $\nB=5000$, in agreement with the measured value in Table~\ref{tab:linearization}. In Section~\ref{sec:drift} the relative error in
$G$ is $1.6\times10^{-5}$ to $4.6\times10^{-4}$ for $0.4\le f\le1$, negligible
against the $O(\nw^{-1/2})$ sampling error. Under drift, however, each batch's mean differs from the window mean by the lag, so $\delta$ moves by an amount set by the drift rate and not by sampling noise. The error reaches several percent for long windows under fast drift (Table~\ref{tab:linearization}). At
the $f$ that is best for a given drift, the penalty is small, at most about $5\%$ in the embedding error.

When all $m$ right directions are tracked, the SKL state $(V,\Lambda)$ is determined
by $G$, and the first-order algorithm obtains it directly from the accumulated statistics $T_0$,
$T_1$ and $c$ by an eigendecomposition, not by SKL rotations of a stored basis. The stored left factor $U$ is never needed, and with it the
row-by-row bookkeeping of the exact method disappears.

The state consists of $T_1$, $T_0$, $c$ and one batch, which is $8(m^3+m^2+3m+2\nB m)$ bytes,
independent of $n$ and of $f$. For $m=50$ and $\nB=5000$ this is 5{,}021{,}200 bytes, that is $5.0$ MB, of which the batch buffers
($16\nB m$ bytes) take 4.0\,MB and $T_1$ ($8m^3$ bytes) 1.0\,MB. This count excludes the $m\times m$ arrays $G$, $V$ and $V_{\mathrm{frame}}$ formed at each step ($24m^2$ bytes, 60\,kB here), just as the equal-memory count for \Rose{} in Section~\ref{sec:experiments} excludes its SVD workspace. Each batch costs $O(\nB m^3)$ to
accumulate $T_1$ and $O(m^3)$ for the eigendecomposition and for re-referencing. Any point is embedded
on demand at cost $O(mk)$ from $(V,\Lambda,\mu)$ alone, so no stored data are needed.

The accuracy depends on the relative degree fluctuation $\varepsilon$, which is small
when each degree averages many landmarks (wide kernels, many landmarks) and larger for narrow
kernels, few landmarks, or sparse regions. Under drift it is also set by the lag, so it grows with the
drift rate and the window length. A second-order term is possible but costs $O(m^4)$ memory and $O(\nB m^4)$
time (50\,MB at $m=50$). For $f<1$ an alternative is to keep the last $K\approx\log(\mathrm{tol})/(2\log f)$ batches, beyond which the weights are
negligible, and form the weighted Gram matrix exactly. For large $m$ the $O(m^3)$ state is the bottleneck (about 1\,GB at $m=500$), and a truncated or
sketched $T_1$ would be needed. Points embedded on arrival use the estimate available at that time. A final re-embedding with the final $(V,\Lambda)$ requires keeping each point's kernel row ($O(m)$ per point)
or its raw coordinates, and under forgetting such a re-embedding describes the stream by the final,
recent-weighted estimate.

\subsection{The Algorithm}\label{sec:alg-words}
Algorithm~\ref{alg:firstorder} maintains, after every batch, the leading
eigenpairs of the age-weighted Gram matrix $G$ of the degree-normalized
landmark affinity, from which any point can be embedded on demand. For each
incoming batch, the algorithm first evaluates the Gaussian kernel block
against the fixed landmarks and forms the batch column sum, exactly as in
\Rose{}. It then decays all accumulated quantities by $f^2$, including the weighted column sum $c$, the weighted count $\nw$, and the moment statistics $T_0$ and $T_1$. The degrees of the new points are
computed against the reference row $\mu_r$, and the batch is added into $T_0$ and
$T_1$ and then discarded. Depending on the forgetting factor, the reference
is moved to the current mean at every batch ($f<1$) or whenever $n$ doubles ($f=1$). The Gram matrix is then assembled from
$T_0$, $T_1$ and $\mu=c/\nw$ as in~\eqref{eq:G1}, symmetrized, and
eigendecomposed with negative eigenvalues clamped to zero, which yields the
tracked right factor $V$ and singular values $\Lambda$. Finally, the
gauge is updated by orthogonal Procrustes, and points are embedded with the
scale-invariant formula~\eqref{eq:embed}. Because we track all $m$ right directions, the SKL residual basis,
core matrix and small SVD reduce to a single eigendecomposition of $G$
and are not needed explicitly. For reference, the exact version of the algorithm that keeps every kernel row is given in Appendix~\ref{sec:appendix}.

\begin{algorithm}[h!]
\caption{First-order \iRose{} (any $f\in[0,1]$)}\label{alg:firstorder}
\begin{algorithmic}[1]
\Require landmarks $Y$; batches $X_0,X_1,\dots$; forgetting factor $f$; $q$, $t$, $\epsilon$
\Ensure tracked $(V,\Lambda)$ after every batch; on-demand embedding $\Phi(x)$
\State $c\gets0$;\ \ $\nw\gets0$;\ \ $T_0\gets0$;\ \ $T_1\gets0$;\ \ $\mu_r\gets$ unset
\For{each batch $X_B$}
  \State $W_B\gets K_\epsilon(X_B,Y)$;\ \ $c_B\gets W_B^\top\one$
  \State $c\gets f^2c+c_B$;\ \ $\nw\gets f^2\nw+\nB$;\ \ $\mu\gets c/\nw$ \Comment{$f$ enters}
  \State $T_0\gets f^2T_0$;\ \ $T_1\gets f^2T_1$
  \If{$\mu_r$ is unset} $\mu_r\gets\mu$;\ \ $n_{\mathrm{ref}}\gets\nw$
  \EndIf
  \State $g\gets W_B\mu_r$ \Comment{reference degrees of the batch}
  \State $T_0\gets T_0+\sum_{i\in B}w_iw_i^\top/g_i$;\ \ $T_1\gets T_1+\sum_{i\in B}w_i\otimes w_i\otimes w_i/g_i^2$
  \If{$f<1$ \textbf{or} $\nw\ge2n_{\mathrm{ref}}$} \Comment{each batch if $f<1$; on doubling if $f=1$}
    \State $T_0\gets T_0-T_1\cdot(\mu-\mu_r)$;\ \ $\mu_r\gets\mu$;\ \ $n_{\mathrm{ref}}\gets\nw$ \Comment{\eqref{eq:rebase}}
  \EndIf
  \State $G\gets\bigl(T_0-T_1\cdot(\mu-\mu_r)\bigr)/\nw$;\ \ $G\gets(G+G^\top)/2$ \Comment{\eqref{eq:G1}}
  \State $[V,\Lambda^2]\gets\eigh(G)$, clamping negative eigenvalues to zero, sorted descending
  \State  update the gauge $\mathcal{G}$ by orthogonal Procrustes against $V_{\mathrm{frame}}$
\EndFor
\State \textbf{embedding:} $\Phi(x)=\bigl(w^\top/\sqrt{w^\top\mu}\bigr)\,V_{2:q+1}\Lambda^{2t-1}_{2:q+1}\,\mathcal{G}$
\end{algorithmic}
\end{algorithm}

\section{Experiments}\label{sec:experiments}

Unless stated otherwise, data points on $\mathbb{S}^1$ are $x_i=(\cos\theta_i,\sin\theta_i)$ with $\theta_i\sim
\mathrm{Unif}[0,2\pi)$. The $m=50$ landmarks are equispaced, $y_k=\bigl(\cos\tfrac{2\pi(k-1)}m,\sin\tfrac{2\pi(k-1)}m\bigr)$.
The kernel is Gaussian, $K_\epsilon(x,y)=e^{-\|x-y\|^2/\epsilon}$ with $\epsilon=0.5$, the embedding dimension is $q=2$ and
the diffusion time is $t=1$. All computations use float64. The drift study (Section~\ref{sec:drift}) uses the same
landmarks and parameters with a non-uniform, rotating density.

On $\mathbb{S}^1$ the Laplace--Beltrami eigenfunctions are $1,\cos j\theta,\sin j\theta$ with eigenvalues
$0,1,1,4,4,9,9,\dots$, so the embedding with $q=2$ should recover the circle $(\cos\theta,\sin\theta)$. At the fixed
bandwidth $\epsilon=0.5$ the reference is not the Laplace--Beltrami limit but the landmark kernel operator itself
(for example $\sigma_2=0.8644$). For uniform sampling and equispaced landmarks this operator is circulant, so its
eigenvectors on the landmarks are exactly the constant and the $\cos j\theta,\sin j\theta$ pairs, and the
eigenvalues are exactly degenerate in pairs. Individual eigenvectors within a pair are therefore not identifiable, and only eigenspaces are. All comparisons below are made after alignment within degenerate pairs, or through rotation-invariant quantities such as subspace distances and Procrustes
errors. We report $k=11$ modes (the constant mode plus five pairs). A choice of $k=10$ would
split the $\cos5\theta/\sin5\theta$ pair.

\begin{table}[t]
\centering
\small
\caption{Exact \iRose{} against plain \Rose{} on $1.6\times10^5$ stationary points. The last two rows compare
with \Rose{} computed directly on the same data with the same age weights $\omega_i=f^{2a_i}$ (for $f=1$ this is
plain \Rose). The rows above the rule compare both columns with unweighted plain \Rose{} on all data, so for $f=0.95$ they measure the effect of the shorter window, not an algorithmic error}.\label{tab:exact}
\begin{tabular}{@{}lcc@{}}
\toprule
Quantity & $f=0.95$ & $f=1$\\
\midrule
$\max_{j\le11}|\sigma_j-\sigma_j^{\mathrm{batch}}|$ & $6.2\times10^{-4}$ & round-off\\
Eigenvector error, modes 1--11 (pair-aligned) & $4.5$--$8.6\times10^{-3}$ & round-off\\
Subspace error, top 11 & $2.8\times10^{-3}$ & round-off\\
$\max(|V^\top V-I|,|U^\top U-I|)$ over the run & $4.8\times10^{-14}$ & $\sim10^{-14}$\\
Embedding RMSE, final re-embedding & $3.8\times10^{-3}$ & $3.2\times10^{-14}$\\
Embedding RMSE, arrival-time, gauge-fixed & $8.3\times10^{-3}$ & $4.1\times10^{-3}$\\
Embedding RMSE, arrival-time, without gauge fixing & 1.08 & ---\\
\midrule
$\max_{j\le11}|\sigma_j-\sigma_j^{\mathrm{wR}}|$, same age weights & $7\times10^{-15}$ & $8\times10^{-14}$\\
Max eigenspace distance, same age weights & $3.5\times10^{-14}$ & $6\times10^{-14}$\\
\bottomrule
\end{tabular}
\end{table}

\subsection{Correctness of the tracked decomposition}\label{sec:correctness}

In the first round of experiments we apply the exact variant of \iRose{} on a data stream consisting of an initial batch of 10{,}000 points followed by 100 batches of 1{,}500, for a total of 160{,}000 points. To check the correctness of the tracked decomposition we compare to plain (unbatched) \Rose{} on the same 160{,}000
points. Embedding errors are orthogonal-Procrustes RMSE against the full, unbatched embedding. We
report the embedding in two ways: \emph{arrival-time}, where each batch is embedded when it arrives and
the results are concatenated, and \emph{final}, where all stored points are re-embedded with the final state
(Algorithm~\ref{alg:exact}, last line).

Table~\ref{tab:exact} summarizes the comparison for $f=1$ and $f=0.95$. For $f=1$ the tracked $(V,\Lambda)$ and the final embedding match plain \Rose{} to round-off,
as Proposition~\ref{prop:exact} predicts. For $f=0.95$ the method computes \Rose{} on age-weighted data, whose effective window is about $\nB/(1-f^2)\approx1.5\times10^4$ points
against all $1.6\times10^5$ for the batch reference. In that case the nonzero differences from plain \Rose{} are the sampling cost of that shorter window
(Section~\ref{sec:forgetting}) and not algorithmic. This is confirmed directly by computing \Rose{} on the same 160{,}000 points with the same age weights, $\Ahat=\Omega^{1/2}D^{-1/2}W$ with $c=W^\top\Omega\one$. With the correct weights, plain \Rose{} agrees with exact \iRose{} to $7\times10^{-15}$ in the singular values and $3.5\times10^{-14}$ in the eigenspaces (last rows of Table~\ref{tab:exact}).

The arrival-time embedding is somewhat less accurate than the final one because each batch is
embedded with the smaller sample available at that time. Without the Procrustes frame, consecutive batches are mutually rotated within the $\cos\theta/\sin\theta$ pair, and the arrival-time RMSE is 1.08. With it, the RMSE is $8.3\times10^{-3}$ at
$f=0.95$ (Table~\ref{tab:exact}). The rotation is applied after the eigenvalue weighting and is an exact isometry,
so it removes the arbitrary rotation without changing the geometry of the embedding.

\subsection{Scaling study} With $f=1$ and full storage, exact \iRose{}
and \Rose{} on the same $n$ points are the same estimator, so there is no accuracy difference
on equal data, as Table~\ref{tab:exact} shows. The advantage of batching must therefore come from resources.
Plain \Rose{} materializes the $n\times m$ matrix $\Ahat$, so its memory grows with $n$, and keeping a
current estimate after every batch costs a full $O(nm^2)$ refit. Exact \iRose{} removes neither
cost, because the degree correction needs all stored rows. The scaling study therefore evaluates
first-order \iRose{} (Algorithm~\ref{alg:firstorder}).

All errors are measured against the population ($n\to\infty$) solution for
the same landmarks, $G_\infty=\mathbb{E}_x\bigl[ww^\top/(w^\top\mu_\infty)\bigr]$, which
we compute by trapezoidal quadrature on a uniform grid of 50{,}000 angles. The integrand is periodic and smooth, so the trapezoidal rule is spectrally
accurate~\cite{trefethen2014}. When the grid size is a multiple of $m$, the matrix $G_\infty$ is circulant, so its
eigenspaces are exactly the constant mode and the $\cos j\theta,\sin j\theta$
pairs on the landmarks. We compare first-order \iRose{} against two versions
of plain \Rose{}. The first runs with unlimited memory on the first $n$
points, up to $9.05\times10^5$. The second runs at equal memory, holding as many points
as fit in the $5.0$ MB state of \iRose{}, which is $n_{\mathrm{cap}}=12{,}553$ points. This
equal-memory count includes only $W$ and ignores the SVD workspace, which
favors plain \Rose{} (with full accounting \iRose{} would overtake plain \Rose{} sooner). Beyond $n_{\mathrm{cap}}$, capped \Rose{} is in effect a fixed-size window of $n_{\mathrm{cap}}$ points, so its accuracy plateaus.

Errors are reported for the leading 11 modes through three metrics. The
singular-value error is $\max_j|\sigma_j-\sigma_j^\infty|$. The eigenfunction
error is $\max_{\gamma}\|\hat P_\gamma-P_\gamma\|_2$ over the population eigenspaces $\gamma$, where $\hat P_\gamma$ and $P_\gamma$ are orthogonal projection matrices onto the tracked and population eigenspaces, respectively. The embedding error is the
relative orthogonal-Procrustes error on 2{,}000 fixed test angles. To summarize
when batching pays off, we define the crossover $n^\ast$ as the first
checkpoint from which the mean error of \iRose{} stays below that of capped
plain \Rose{} by more than one combined standard deviation, computed over the 5 seeds. All runs use a batch
size of 5{,}000 and extend up to $30\times10^6$ points.

\begin{table}[h!]
\centering
\small
\caption{Stationary scaling study, $f=1$: mean error (5 seeds) against the population solution.
``Capped'' is plain \Rose{} at equal memory ($n_{\mathrm{cap}}=12{,}553$).}\label{tab:scaling}
\begin{tabular}{@{}rcccccc@{}}
\toprule
 & \multicolumn{2}{c}{Singular values} & \multicolumn{2}{c}{Eigenfunctions} & \multicolumn{2}{c}{Embedding}\\
\cmidrule(lr){2-3}\cmidrule(lr){4-5}\cmidrule(lr){6-7}
$n$ & \iRose & capped & \iRose & capped & \iRose & capped\\
\midrule
5{,}000 & $5.1\times10^{-3}$ & $5.1\times10^{-3}$ & $2.8\times10^{-2}$ & $2.8\times10^{-2}$ & $2.2\times10^{-2}$ & $2.2\times10^{-2}$\\
40{,}000 & $1.4\times10^{-3}$ & $2.6\times10^{-3}$ & $9.1\times10^{-3}$ & $1.5\times10^{-2}$ & $6.7\times10^{-3}$ & $9.5\times10^{-3}$\\
80{,}000 & $9.4\times10^{-4}$ & $2.6\times10^{-3}$ & $5.3\times10^{-3}$ & $1.5\times10^{-2}$ & $4.4\times10^{-3}$ & $9.5\times10^{-3}$\\
905{,}000 & $2.3\times10^{-4}$ & $2.6\times10^{-3}$ & $1.6\times10^{-3}$ & $1.5\times10^{-2}$ & $1.2\times10^{-3}$ & $9.5\times10^{-3}$\\
2{,}000{,}000 & $2.0\times10^{-4}$ & $2.6\times10^{-3}$ & $1.2\times10^{-3}$ & $1.5\times10^{-2}$ & $7.6\times10^{-4}$ & $9.5\times10^{-3}$\\
10{,}000{,}000 & $8.8\times10^{-5}$ & $2.6\times10^{-3}$ & $6.2\times10^{-4}$ & $1.5\times10^{-2}$ & $5.5\times10^{-4}$ & $9.5\times10^{-3}$\\
30{,}000{,}000 & $5.9\times10^{-5}$ & $2.6\times10^{-3}$ & $3.5\times10^{-4}$ & $1.5\times10^{-2}$ & $3.1\times10^{-4}$ & $9.5\times10^{-3}$\\
\bottomrule
\end{tabular}
\end{table}

Table~\ref{tab:scaling} gives the mean error against the population solution.
Plain \Rose{} with unlimited memory matches \iRose{} at every $n$ it reaches (up to
$9.05\times10^5$), to within the linearization error discussed below. Above $n_{\mathrm{cap}}$, the error of capped plain \Rose{}
is frozen at its $n_{\mathrm{cap}}$ value, as expected of a fixed window, while the error of \iRose{} keeps
decreasing. A log--log fit gives slopes of $-0.52$ (singular values), $-0.52$ (eigenfunctions) and $-0.54$
(embedding), consistent with the $n^{-1/2}$ rate of sampling error. The linearization error, measured
as the difference between \iRose{} and plain \Rose{} on the same data, is at most $6\times10^{-4}$ (eigenfunctions)
for $n\le1.5\times10^4$ and falls to $8\times10^{-6}$ (embedding) and $4\times10^{-6}$ (singular values) by $9\times10^5$.
Its log--log slopes over $10^4\le n\le9\times10^5$ are $-0.77$ (singular values), $-0.84$ (eigenfunctions) and $-0.82$
(embedding). Relative to the sampling error it is 0.5--3\% for the eigenfunctions and the embedding and
1.5--5\% for the singular values, largest just after the first re-reference ($n=1$--$1.5\times10^4$) and decreasing with $n$.

\begin{table}[h!]
\centering
\small
\caption{Summary of scaling-study findings (5 seeds, batch size 5{,}000, $f=1$).}\label{tab:scaling-summary}
\begin{tabular}{@{}lccc@{}}
\toprule
 & Singular values & Eigenfunctions & Embedding\\
\midrule
Error decay, log--log slope & $-0.52$ & $-0.52$ & $-0.54$\\
Linearization error, log--log slope ($10^4\le n\le9\times10^5$) & $-0.77$ & $-0.84$ & $-0.82$\\
Linearization error / sampling error & 1.5--5\% & 0.5--3\% & 0.7--3\%\\
Accuracy crossover $n^\ast$ at equal memory & 80{,}000 & 40{,}000 & 55{,}000\\
Error ratio, capped plain / \iRose, at $2\times10^6$ & 13.0$\times$ & 13.0$\times$$^{\dagger}$ & 12.5$\times$\\
Error ratio, capped plain / \iRose, at $30\times10^6$ & 43.4$\times$ & 42.9$\times$ & 31$\times$\\
\bottomrule
\end{tabular}
\end{table}

\begin{figure}[h!]
\centering
\figfile[width=0.95\linewidth]{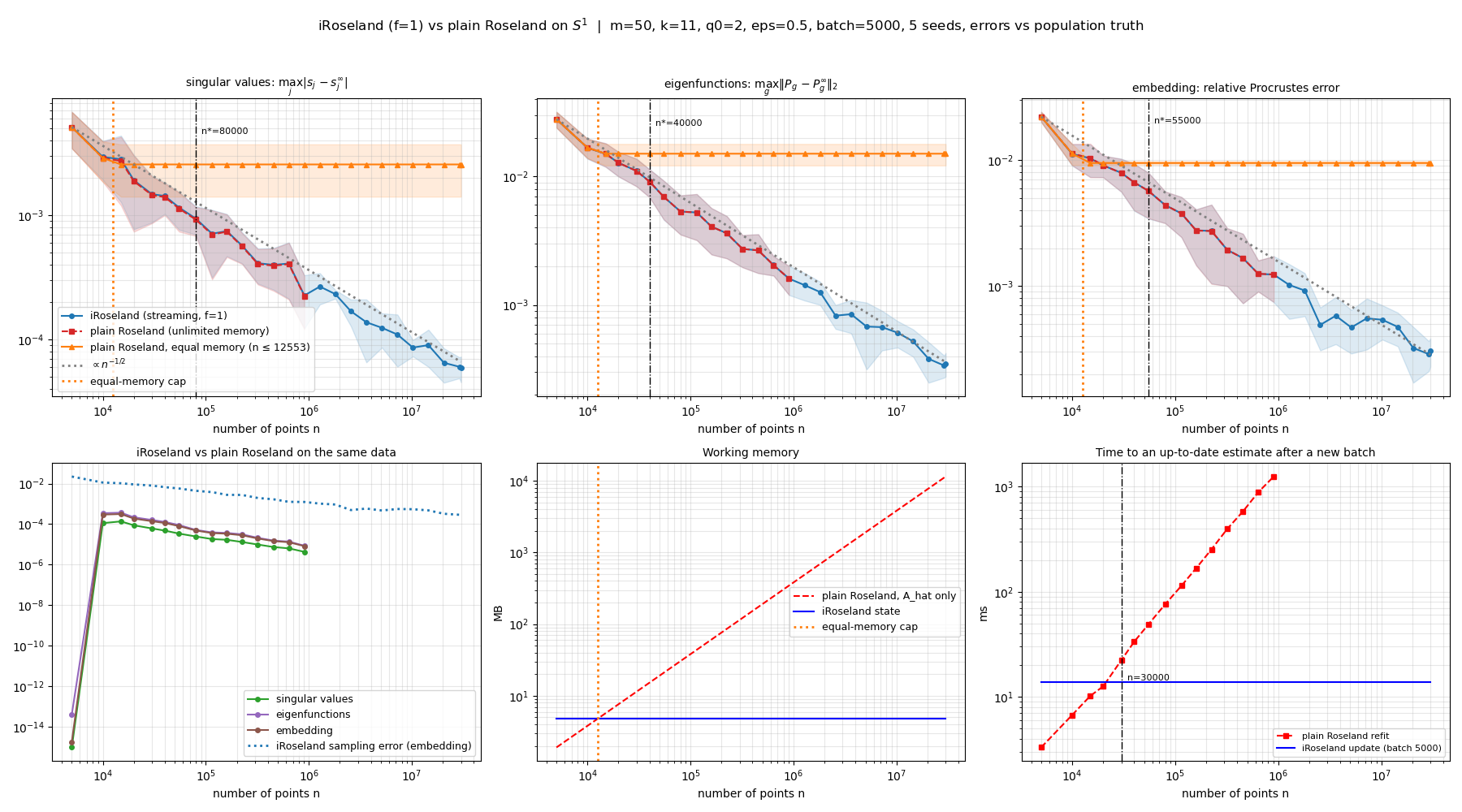}
\caption{Stationary scaling study. Error curves for \iRose, unlimited-memory plain \Rose{}
and capped plain \Rose; linearization error; memory; and update latency.}\label{fig:scaling}
\end{figure}

The state of \iRose{} is $5.0$ MB for $m=50$ and $\nB=5{,}000$, independent of $n$. For a memory budget $M$ and $m$ landmarks, plain \Rose{} holds $n_{\mathrm{cap}}\approx M/(8m)$
points, while \iRose{} needs $8(m^3+m^2+3m+2\nB m)$ bytes, that is $O(m^3+\nB m)$. At these settings the batch
buffers ($16\nB m$ bytes, 4.0\,MB) dominate $T_1$ ($8m^3$ bytes, 1.0\,MB), and $8m^3$ dominates only when $m^2\gtrsim2\nB$. The crossover therefore sits a few multiples
above $n_{\mathrm{cap}}$. Beyond $n_{\mathrm{cap}}$ the error ratio of capped \Rose{} to \iRose{} grows like $\sqrt{n/n_{\mathrm{cap}}}$, and the crossover occurs once this ratio exceeds one by more than the relative run-to-run standard deviation of the two errors. This happens at about $n^\ast\approx(3$--$6.5)\,n_{\mathrm{cap}}$ here.
\iRose{} is preferable whenever the expected stream length exceeds a few times $m^2+2\nB$. An
\iRose{} update takes 39.7\,ms per batch of 5{,}000 points, constant in $n$. A plain \Rose{} refit
to keep a current estimate takes 8\,ms at $5\times10^3$ points, 319\,ms at $1.6\times10^5$ and 2.1\,s at $9.05\times10^5$.
The latency crossover is at $n\approx40{,}000$.

\begin{figure}
    \centering
    \includegraphics[width=\textwidth]{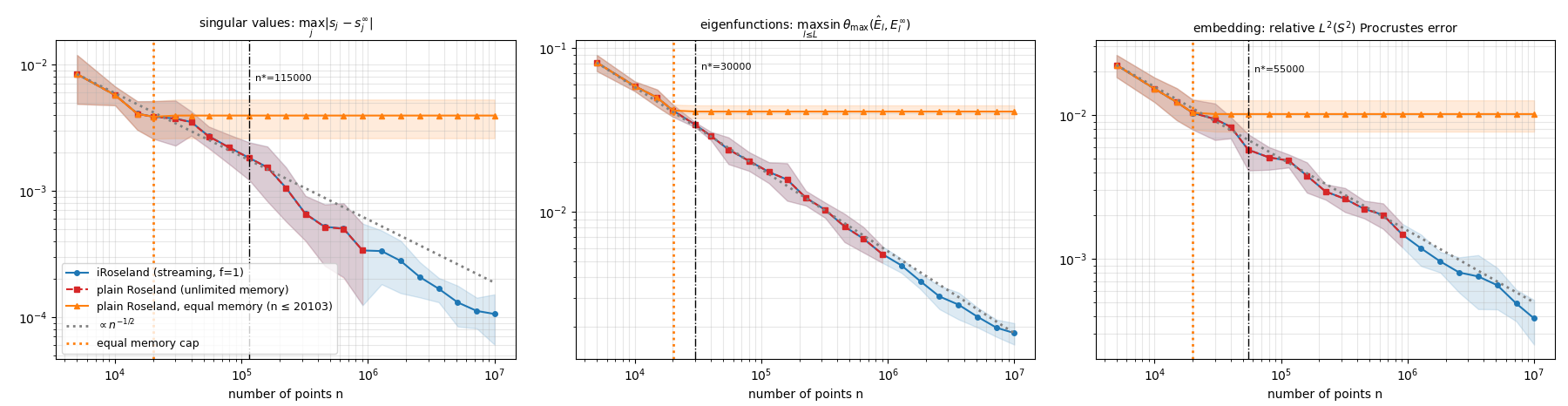}
    \caption{Stationary ($f=1$) scaling study on the sphere, up to $10^7$ points. Error against the population solution for the singular values (left), eigenfunctions (center), and embedding (right).}
    \label{fig:scaling-sphere}
\end{figure}

We repeat the stationary scaling study on $\mathbb{S}^2$ with $f=1$, using the same
protocol as on $\mathbb{S}^1$ ($5$ seeds, errors against the
$n\to\infty$ population solution), and extend it to $n = 10^7$ points. The results closely mirror those of the circle (Figure~\ref{fig:scaling-sphere}).
Plain Roseland with unlimited memory is matched by iRoseland at every $n$
for which it was computed (up to $n=10^6$), to within the linearization error. \iRose{} begins to show improvement over memory capped \Rose{} beyond $n_{\mathrm{cap}}$, and continues to improve as more 
points are processed.

The conclusions of the scaling study apply to the $f=1$ regime. With $f<1$ the error of first-order \iRose{} would level off at the noise floor of an effective sample
of about $\ness=(1+f^2)\nw$ points, much like the memory-capped plain \Rose.

\subsection{Drifting data}\label{sec:drift}

For drifting data we use first-order \iRose{} (Algorithm~\ref{alg:firstorder}), with the reference moved at every batch
for $f<1$ and at every batch for $f=1$ as well, so that comparisons are fair (the doubling rule of
Algorithm~\ref{alg:firstorder}, line~10, is not used in this study). Here points on $\mathbb{S}^1$ follow
a von Mises density with concentration $\kappa=1$ whose peak is at $\Delta\tau$ in batch $\tau=0,1,\dots,199$, so it rotates by $\Delta$ radians per batch,
$\Delta\in\{0,0.002,0.01,0.05\}$. At $\Delta=0.05$ the total rotation is 10\,rad, more than one revolution. There
are 200 batches of $\nB=5{,}000$ points. Errors are measured against the $n\to\infty$ \Rose{} solution for
the \emph{current} density, computed by quadrature on a grid of 20{,}000 angles weighted by the von Mises density, and averaged
over the second half of the stream at every tenth batch. Because the density is not uniform, the eigenvalue pairs are no longer degenerate, and the metrics differ from those of the scaling study. Here we consider singular values ($\max_{j\le5}|\sigma_j-\sigma_j^\infty|$), eigenspace (the subspace distance between the leading three right
singular vectors $V_{:,1:3}$ and their population counterparts), and embedding (the relative orthogonal-Procrustes error on 2{,}000
uniformly spaced test angles). Table~\ref{tab:drift} uses one seed, reused for every $f$,
with $f\in\{1,0.995,0.99,0.98,0.95,0.9,0.8\}$ (Table~\ref{tab:drift} shows a subset). To separate the forgetting trade-off from the linearization, we
also compute \Rose{} on the same age-weighted data directly (Table~\ref{tab:drift-exact}, and compare the two on the same stream for selected $(\Delta,f)$
(Table~\ref{tab:linearization}).

\begin{table}[h!]
\centering
\small
\caption{Drifting data: embedding error of first-order \iRose{} against the $n\to\infty$ solution for the current density (von
Mises, $\kappa=1$; 200 batches of 5{,}000; one seed). The bold entry is the smallest in each row. The window
$\nw\approx\nB/(1-f^2)$ is shown for each $f$ at the end of the stream; for $f=1$ it is the total of $10^6$ points, and over the
averaging window it ranges from $5.5\times10^5$ to $10^6$ (from $2.2\times10^5$ to $2.5\times10^5$ for $f=0.99$).}\label{tab:drift}
\begin{tabular}{@{}lccccc@{}}
\toprule
Drift per batch & $f=1$ & $f=0.99$ & $f=0.95$ & $f=0.9$ & $f=0.8$\\
Window $\nw$ & $10^6$ & $2.5\times10^5$ & $5.1\times10^4$ & $2.6\times10^4$ & $1.4\times10^4$\\
\midrule
0 (stationary) & $\mathbf{1.1\times10^{-3}}$ & $1.8\times10^{-3}$ & $4.5\times10^{-3}$ & $6.2\times10^{-3}$ & $8.7\times10^{-3}$\\
0.002 rad & $6.7\times10^{-2}$ & $3.6\times10^{-2}$ & $8.8\times10^{-3}$ & $\mathbf{7.2\times10^{-3}}$ & $8.9\times10^{-3}$\\
0.01 rad & $3.4\times10^{-1}$ & $2.0\times10^{-1}$ & $4.1\times10^{-2}$ & $2.0\times10^{-2}$ & $\mathbf{1.2\times10^{-2}}$\\
0.05 rad & $4.7\times10^{-1}$ & $4.5\times10^{-1}$ & $2.3\times10^{-1}$ & $1.0\times10^{-1}$ & $\mathbf{4.2\times10^{-2}}$\\
\bottomrule
\end{tabular}
\end{table}

Table~\ref{tab:bestf} gives the best $f$ for each metric for the five-seed age-weighted study. The singular-value and eigenspace errors do not follow
the embedding exactly, since they prefer longer windows. The overall pattern
is the trade-off of Section~\ref{sec:forgetting}. With no drift, the best
choice is $f=1$ in every metric, and the error increases as the window
shrinks. Even very slow drift ($0.002$\,rad per batch, $0.4$\,rad in total)
makes $f=1$ about 9 times less accurate in the embedding than the best $f$
($6.7\times10^{-2}$ against $7.5\times10^{-3}$ at $f=0.9$, Table~\ref{tab:drift-exact}) and about
3 times less accurate in the leading eigenspace. The singular values are not degraded at this drift rate because a rotation of the density leaves them invariant and only the broadening of
Section~\ref{sec:forgetting} affects them. The faster the drift, the smaller the
best $f$ (Table~\ref{tab:bestf}). The extended grid of Table~\ref{tab:drift-exact} brackets every optimum, and using only the latest
batch ($f=0$) is never best. The best $f$
depends on the metric, ordered from the longest window to the shortest as
singular values, eigenspace and embedding, as expected from the different
sensitivity of the metrics to the shift and the broadening.

\begin{table}[h!]
\centering
\small
\caption{Drifting data: embedding error (in units of $10^{-3}$) of \Rose{} on the same age-weighted data, computed directly,
against the $n\to\infty$ solution for the current density (protocol of Table~\ref{tab:drift}; mean over 5 seeds). The bold entry is the
smallest in each row. Here $f=0$
uses only the latest batch.}\label{tab:drift-exact}
\setlength{\tabcolsep}{4.5pt}
\begin{tabular}{@{}lcccccccccc@{}}
\toprule
Drift per batch & $f=1$ & 0.99 & 0.95 & 0.9 & 0.8 & 0.7 & 0.6 & 0.5 & 0.4 & 0\\
\midrule
0 (stationary) & \textbf{1.6} & 2.0 & 4.3 & 6.3 & 9.0 & 11.1 & 12.8 & 14.3 & 15.6 & 18.0\\
0.002 rad & 67.2 & 36.2 & 9.2 & \textbf{7.5} & 9.2 & 11.1 & 12.8 & 14.3 & 15.6 & 18.0\\
0.01 rad & 320 & 174 & 40.6 & 19.8 & 12.1 & \textbf{12.0} & 13.1 & 14.4 & 15.6 & 18.0\\
0.05 rad & 478 & 437 & 182 & 91.0 & 39.9 & 23.7 & 17.7 & \textbf{16.0} & 16.1 & 18.0\\
\bottomrule
\end{tabular}
\end{table}

\begin{table}[h!]
\centering
\small
\caption{Drifting data: best $f$ for each metric (age-weighted \Rose, 5 seeds).}\label{tab:bestf}
\begin{tabular}{@{}lccc@{}}
\toprule
Drift per batch & Singular values & Eigenspace & Embedding\\
\midrule
0 & 1 & 1 & 1\\
0.002 rad & 0.99 ($\approx1$) & 0.95 & 0.9\\
0.01 rad & 0.95 & 0.9 & 0.7 (0.7--0.8)\\
0.05 rad & 0.8 & 0.7 & 0.5 (0.4--0.6)\\
\bottomrule
\end{tabular}
\end{table}

Table~\ref{tab:linearization} compares first-order \iRose{} with age-weighted \Rose{} on the same
stream. On stationary data the two agree to a relative error in $G$ of $1.6\times10^{-5}$ to $4.6\times10^{-4}$, so the errors in
Table~\ref{tab:drift} are those of the age-weighted estimator. Under drift the linearization error grows roughly as $\Delta^2/(1-f^2)^2$
(driven mainly by the stale $T_1$): at 0.05 rad per batch and $f=0.95$ the relative error in $G$ is 9\%, the embedding error is $0.23$
against $0.18$ for the exact weighting, and the singular-value error is 2.7 times larger. At the $f$ that suits the drift, the
penalty is small: at 0.01 rad and $f=0.8$ the embedding errors agree to three digits, and at 0.05 rad the embedding error is 5\% larger at
$f=0.8$ and 2\% larger at $f=0.6$. The entries of Table~\ref{tab:drift} for long windows under fast drift therefore combine lag error and
linearization error.

\begin{table}[h!]
\centering
\tiny
\caption{Drifting data: first-order \iRose{} ($G_1$) against \Rose{} on the same age-weighted data ($G_x$), same stream
(seed 1). Errors against the $n\to\infty$ solution for the current density, averaged as in Table~\ref{tab:drift}.}\label{tab:linearization}
\begin{tabular}{@{}lcccc@{}}
\toprule
Drift per batch & $f$ & $\|G_1-G_x\|_F/\|G_x\|_F$ & Embedding error, first order / exact & Singular-value error, first order / exact\\
\midrule
0 & 0.95 & $2\times10^{-5}$ & $4.50\times10^{-3}$ / $4.50\times10^{-3}$ & $1.15\times10^{-3}$ / $1.15\times10^{-3}$\\
0 & 0.4 & $5\times10^{-4}$ & $1.66\times10^{-2}$ / $1.66\times10^{-2}$ & $5.44\times10^{-3}$ / $5.39\times10^{-3}$\\
0.01 rad & 0.95 & $5.4\times10^{-3}$ & $4.10\times10^{-2}$ / $3.99\times10^{-2}$ & $2.69\times10^{-3}$ / $1.41\times10^{-3}$\\
0.01 rad & 0.8 & $3.8\times10^{-4}$ & $1.19\times10^{-2}$ / $1.19\times10^{-2}$ & $2.76\times10^{-3}$ / $2.72\times10^{-3}$\\
0.05 rad & 0.95 & $9.0\times10^{-2}$ & $2.33\times10^{-1}$ / $1.81\times10^{-1}$ & $3.67\times10^{-2}$ / $1.35\times10^{-2}$\\
0.05 rad & 0.9 & $3.1\times10^{-2}$ & $1.03\times10^{-1}$ / $9.05\times10^{-2}$ & $1.17\times10^{-2}$ / $4.7\times10^{-3}$\\
0.05 rad & 0.8 & $8.5\times10^{-3}$ & $4.20\times10^{-2}$ / $3.98\times10^{-2}$ & $5.0\times10^{-3}$ / $3.0\times10^{-3}$\\
0.05 rad & 0.6 & $2.4\times10^{-3}$ & $1.85\times10^{-2}$ / $1.81\times10^{-2}$ & $4.6\times10^{-3}$ / $4.2\times10^{-3}$\\
\bottomrule
\end{tabular}
\end{table}

\bibliographystyle{unsrt}
\bibliography{references}

\appendix
\section{Exact Degree Correction}
\label{sec:appendix}

When a batch arrives, the stored degrees are updated as
$d\leftarrow f^2d_{\mathrm{old}}+Wc_B$, and each old row of $\hat A$ is
rescaled by $s_i=\sqrt{d_i^{\mathrm{old}}/d_i}$. Because $d_i\ge f^2d_i^{\mathrm{old}}$,
the combined factor satisfies $fs_i\le1$, although $s_i$ alone may exceed one when $f<1$. Both factors are computed from the same age weights, as required for the result to be \Rose{} on age-weighted data. The rescaled old block
$\mathrm{diag}(s)\,U\Lambda V^\top$ is then refactorized in two steps. First, re-orthogonalize the rescaled left factor, $\diag(s)\,U=Q_UR_{\mathrm{qr}}$, with $Q_U$ having orthonormal columns. Then
\[\diag(s)\,U\Lambda V^\top=Q_U\,(R_{\mathrm{qr}}\Lambda)\,V^\top.\]
The core $R_{\mathrm{qr}}\Lambda$ is upper triangular, not diagonal, because the non-uniform scaling $s$ mixes the modes. Replacing it by its diagonal discards this mixing and fixes $V$, which is not a valid refactorization. The exact step takes the SVD of the small $k_t\times k_t$ core,
\[R_{\mathrm{qr}}\Lambda=P\,\Lambda_c\,\tilde Q^\top,\qquad U\gets Q_UP,\quad V\gets V\tilde Q,\quad\Lambda\gets\Lambda_c,\]
after which $U\Lambda V^\top=\diag(s)\,U_{\mathrm{old}}\Lambda_{\mathrm{old}}V_{\mathrm{old}}^\top$ exactly, with orthonormal $U$ and $V$. The cost is a QR of an $n\times k_t$ matrix per batch, $O(nk_t^2)$. The first-order alternative
replaces this step with an expansion of the degrees in aggregate statistics, as
described in Section~\ref{sec:firstorder}. 

Two details are needed for the SKL step to remain exact once the degree
correction revisits stored rows. The first concerns the residual basis $V_p$ in
Equation~\eqref{eq:skl-rose}, which must be an orthonormal basis of the part of
$Q_B$ orthogonal to $V$, as in Ross et al. A residual $Q_B-VV^\top Q_B$ has rank at most $m-k_t$, whereas a plain economy QR of an $m\times n_B$ matrix with $n_B\ge m$ returns a full $m\times m$ orthogonal matrix that spans $\R^m$ and therefore overlaps $\operatorname{range}(V)$. The core matrix then double-counts the in-subspace part of every batch and $[V\ \ V_p]$ is no longer orthonormal. We therefore compute $V_p$ from a rank-revealing SVD of the residual, projected against $V$ twice, keeping the left singular vectors whose singular value exceeds $\tau\|Q_B\|_F$ (we use $\tau=10^{-12}$), orthogonalizing once more against $V$, and finishing with a QR. The tolerance is relative to $\|Q_B\|_F$, not to the largest residual singular value, so that a numerically zero residual yields $r=0$. Because $V_p\perp V$, the lower-right block of $R$ can then be written as $V_p^\top Q_B$, which equals $V_p^\top(Q_B-VV^\top Q_B)$.

The second detail concerns the stored left factor, which must be carried along.
Ross et al.\ discard $Z_R$ because they never revisit old observations. Here
the degree correction revisits the stored $U$ at every update, so $U$ is
updated as $U\leftarrow\begin{bmatrix}U&0\\0&I\end{bmatrix}Z_R$. The old block
is rotated by the first block of rows of $Z_R$, while the remaining rows give
the left vectors of the new points, which coincide with the out-of-sample
formula in Equation~\eqref{eq:oos} evaluated with the updated $V$ and $\Lambda$.
If the old block were left unrotated, $U$ would lose orthonormality and the
next QR factorization in the degree correction would count the previous batch's
energy a second time.

Exactness also requires that no singular value be discarded before the
correction acts. Truncating to $k_t<m$ at each SKL step would leave the degree
correction acting on a truncated $U\Lambda V^\top$, so the identity could not be
maintained. Since the right space is only $m$-dimensional, tracking all $m$
triplets is inexpensive for moderate $m$. In that case $V_p$ is empty and no
singular value is lost, and the leading $k$ modes are reported. When $m$ is
large and $n$ is larger still, truncation to $k_t<m$ is unavoidable and the
orthonormal residual basis becomes essential, and the guarantees stated below
then hold only approximately. Each batch is embedded as described in
Section~\ref{subsec:embedding}, using the sample-size-invariant scale and the
Procrustes gauge update given there.

\begin{algorithm}[h!]
\caption{Exact \iRose{} (any $f\in[0,1]$)}\label{alg:exact}
\begin{algorithmic}[1]
\Require landmarks $Y$; initial batch $X_0$; batches $X_1,X_2,\dots$; forgetting factor $f$ ($f=1$ for stationary data, $f<1$ for drifting data); $q$, $t$, $\epsilon$; tracked rank $\kt=m$
\Ensure tracked $(V,\Lambda)$; arrival-time embedding $\Phi$; final embedding $\Phi_{\mathrm{final}}$
\State $W\gets K_\epsilon(X_0,Y)$;\ \ $c\gets W^\top\one$;\ \ $d\gets Wc$ 
\State $[U,\Lambda,V]\gets\svd\bigl(\diag(d)^{-1/2}W\bigr)$, keeping $\kt=m$ triplets 
\State $\nw\gets|X_0|$;\ \ $\mathcal{G}\gets I_q$;\ \ $V_{\mathrm{frame}}\gets V_{2:q+1}$
\State $\Phi\gets\textsc{Embed}(W,d)$
\For{each batch $X_B$}
  \State $W_B\gets K_\epsilon(X_B,Y)$;\ \ $c_B\gets W_B^\top\one$ 
  \State $d^{\mathrm{old}}\gets d$;\ \ $d\gets f^2d^{\mathrm{old}}+Wc_B$;\ \ $s\gets\sqrt{d^{\mathrm{old}}/d}$ \Comment{$f$ enters}
  \State $[Q_U,R_{\mathrm{qr}}]\gets\qr(\diag(s)\,U)$
  \State \hspace{2.2em}$[P,\Lambda_c,\tilde Q]\gets\svd(R_{\mathrm{qr}}\Lambda)$;\ \ $U\gets Q_UP$;\ \ $V\gets V\tilde Q$
  \State $\Lambda_f\gets f\Lambda_c$;\ \ $\nw\gets f^2\nw+\nB$
  \State $c\gets f^2c+c_B$;\ \ $d_B\gets W_Bc$;\ \ $\Bhat\gets\diag(d_B)^{-1/2}W_B$
  \State $[V,\Lambda,Z_R]\gets\textsc{SKL}(V,\Lambda_f,\Bhat,\kt)$ 
  \State $U\gets\begin{bmatrix}U\,(Z_R)_{1:\kt,:}\\(Z_R)_{\kt+1:,:}\end{bmatrix}$;\ \ $W\gets\begin{bmatrix}W\\W_B\end{bmatrix}$;\ \ $d\gets\begin{bmatrix}d\\d_B\end{bmatrix}$ 
  \State $[U_{\mathcal{G}},\sim,V_{\mathcal{G}}]\gets\svd(V_{2:q+1}^\top V_{\mathrm{frame}})$;\ \ $\mathcal{G}\gets U_{\mathcal{G}}V_{\mathcal{G}}^\top$;\ \ $V_{\mathrm{frame}}\gets V_{2:q+1}\mathcal{G}$
  \State \hspace{2.2em}$\Phi\gets\begin{bmatrix}\Phi\\\textsc{Embed}(W_B,d_B)\end{bmatrix}$
\EndFor
\State $\Phi_{\mathrm{final}}\gets\textsc{Embed}(W,d)$ \Comment{optional re-embedding}
\end{algorithmic}
\end{algorithm}

\subsection{Exactness}

\begin{proposition}[Exact tracking]\label{prop:exact}
Assume $\kt=m$ and $n_{\mathrm{init}}\ge m$, let the kernel be entrywise positive, and work in exact arithmetic.
After every update of Algorithm~\ref{alg:exact}, with any $f\in[0,1]$, $U\Lambda V^\top$ equals the weighted \Rose{}
matrix~\eqref{eq:weighted} of all data seen so far, with $\omega_i=f^{2a_i}$, and $U$ and $V$ have orthonormal columns. In
particular, for $f=1$ the tracked $(V,\Lambda)$ are those of plain \Rose{} on all data, and for every $f$
the tracked $\sigma_1$ equals one.
\end{proposition}
\begin{proof}
We argue by induction on the number of updates. The invariant is that after
the $j$th update, the tracked triple satisfies
\begin{equation}\label{eq:invariant}
U\Lambda V^\top=\hat A_j,\qquad U^\top U=I,\qquad V^\top V=I,
\end{equation}
where $\hat A_j=\Omega^{1/2}D^{-1/2}W$ is the weighted matrix in
Equation~\eqref{eq:weighted} built from all data seen so far, with weights
$\omega_i=f^{2a_i}$, where $a_i$ is the number of updates since $x_i$ arrived.

\emph{Base case.} The initial batch has $n_{\mathrm{init}}\ge m$ points and every
weight equals one, so $\hat A_0=D^{-1/2}W$. The initialization computes its thin
SVD and keeps all $m$ triplets, so \eqref{eq:invariant} holds, and no singular
value is discarded.

\emph{Inductive step.} Suppose \eqref{eq:invariant} holds before a batch $X_B$
arrives, so that $U\Lambda V^\top=\hat A_{\mathrm{old}}$. We first record how the
target matrix changes. All old weights are multiplied by $f^2$ and the new
points receive weight one, so the weighted column sum and degrees become
\[
c\leftarrow f^2c_{\mathrm{old}}+c_B,\qquad
d_i=f^2d_i^{\mathrm{old}}+w_i^\top c_B\quad\text{for old points},
\]
and the new points have degrees $d_B=W_Bc$ computed with the updated $c$.
Each old row of $\hat A$ therefore changes from
$\sqrt{\omega_i^{\mathrm{old}}/d_i^{\mathrm{old}}}\,w_i$ to
$\sqrt{f^2\omega_i^{\mathrm{old}}/d_i}\,w_i$, which equals
$f\,s_i$ times the old row with $s_i=\sqrt{d_i^{\mathrm{old}}/d_i}$. The new
target matrix is thus
\begin{equation}\label{eq:target-update}
\hat A_{\mathrm{new}}=
\begin{bmatrix} f\,\mathrm{diag}(s)\;U\Lambda V^\top\\ \hat A_B\end{bmatrix},
\qquad \hat A_B=D_B^{-1/2}W_B,
\end{equation}
and we show that the update produces an exact SVD of this matrix.

The thin QR factorization
$\mathrm{diag}(s)\,U=Q_UR_{\mathrm{qr}}$ is exact, with $Q_U$ having orthonormal
columns, so $\mathrm{diag}(s)\,U\Lambda V^\top=Q_U\,R_{\mathrm{qr}}\Lambda\,V^\top$.
The SVD of the small core, $R_{\mathrm{qr}}\Lambda=P\Lambda_c\tilde Q^\top$, gives
$\mathrm{diag}(s)\,U\Lambda V^\top=U'\Lambda_c V'^\top$ with $U'=Q_UP$ and
$V'=V\tilde Q$. Both $U'$ and $V'$ have orthonormal columns, being products of
matrices with orthonormal columns and orthogonal factors. Multiplying by $f$
then gives $f\,\mathrm{diag}(s)\,U\Lambda V^\top=U'\,(f\Lambda_c)\,V'^\top$, which
is exactly the first block of \eqref{eq:target-update}.

Because $c$ and $d_B$ have been updated to the
weights above, $\hat A_B$ is by construction exactly the new block of
$\hat A_{\mathrm{new}}$, with weight one.

Transposing
\eqref{eq:target-update} gives the form of Equation~\eqref{eq:skl-rose},
\[
\bigl[\,V'f\Lambda_cU'^\top\;\;\hat A_B^\top\,\bigr]
=\bigl[\,V'\;\;V_p\,\bigr]\,R\,
\begin{bmatrix}U'&0\\0&I\end{bmatrix}^{\!\top},
\qquad
R=\begin{bmatrix}f\Lambda_c&V'^\top\hat A_B^\top\\0&V_p^\top\hat A_B^\top\end{bmatrix},
\]
where $V_p$ is an orthonormal basis of the part of $\hat A_B^\top$ orthogonal to
$V'$. By construction the matrix $[V'\;V_p]$ has orthonormal columns, and when
$k_t=m$ the matrix $V'$ is already orthogonal, so $V_p$ is empty and the
statement is trivial. The block matrix on the right also has orthonormal
columns. Hence the singular values and right and left singular vectors of the
stacked matrix are obtained exactly from the SVD of the small matrix $R$,
$R=U_R\Sigma_R Z_R^\top$.

The updated right factor
$V_{\mathrm{new}}=[V'\;V_p]\,U_R$ lives in $\R^m$, so $\hat A_{\mathrm{new}}$
has at most $m$ nonzero singular values. Since we keep $k_t=m$ triplets, no
nonzero singular value is discarded.

The left factor is updated as
$U_{\mathrm{new}}=\begin{bmatrix}U'&0\\0&I\end{bmatrix}Z_R$. This is a product of two matrices with orthonormal columns, so
$U_{\mathrm{new}}^\top U_{\mathrm{new}}=I$. The old block is rotated, and the
new points receive their own left vectors.

Together these steps give $U_{\mathrm{new}}\Sigma_RV_{\mathrm{new}}^\top=\hat A_{\mathrm{new}}$
with orthonormal factors, which is \eqref{eq:invariant} for the next update, and
induction completes the proof of the first claim.

For the remaining claims, note that when $f=1$ all weights equal one and
$\hat A=D^{-1/2}W$ is the matrix used by plain \Rose{} on all data seen so far.
An SVD is unique up to signs and rotations within repeated singular values, so
the tracked singular values and singular subspaces coincide with those of
plain \Rose{}. For every $f$, the tracked matrix is the weighted
$\hat A$ of Equation~\eqref{eq:weighted}, and Lemma~\ref{lem:trivial} gives
$\sigma_1=1$ for any positive diagonal weighting.
\end{proof}

\end{document}